\documentclass{article}

\PassOptionsToPackage{numbers, compress}{natbib}

\usepackage[preprint]{nips_2018}

\usepackage[utf8]{inputenc} 
\usepackage[T1]{fontenc}    
\usepackage{hyperref}       
\usepackage{url}            
\usepackage{booktabs}       
\usepackage{amsfonts}       
\usepackage{nicefrac}       
\usepackage{microtype}      
\usepackage{array}
\usepackage{amsmath}
\usepackage{graphicx}
\usepackage{wrapfig}

\title{ChannelNets: Compact and Efficient Convolutional Neural Networks via Channel-Wise Convolutions}

\author{
  Hongyang Gao \\
  Texas A\&M University\\
  College Station, TX 77843 \\
  \texttt{hongyang.gao@tamu.edu} \\
  \And
  Zhengyang Wang \\
  Texas A\&M University\\
  College Station, TX 77843 \\
  \texttt{zhengyang.wang@tamu.edu} \\
  \And
  Shuiwang Ji \\
  Texas A\&M University\\
  College Station, TX 77843 \\
  \texttt{sji@tamu.edu} \\
}

\begin{document}

\maketitle

\begin{abstract}

Convolutional neural networks~(CNNs) have shown great capability of
solving various artificial intelligence tasks. However, the
increasing model size has raised challenges in employing them in
resource-limited applications. In this work, we propose to compress
deep models by using channel-wise convolutions, which replace dense
connections among feature maps with sparse ones in CNNs. Based on
this novel operation, we build light-weight CNNs known as
ChannelNets. ChannelNets use three instances of channel-wise
convolutions; namely group channel-wise convolutions, depth-wise
separable channel-wise convolutions, and the convolutional
classification layer. Compared to prior CNNs designed for mobile
devices, ChannelNets achieve a significant reduction in terms of the
number of parameters and computational cost without loss in
accuracy. Notably, our work represents the first attempt to compress
the fully-connected classification layer, which usually accounts for
about $25$\% of total parameters in compact CNNs. Experimental
results on the ImageNet dataset demonstrate that ChannelNets achieve
consistently better performance compared to prior methods.

\end{abstract}

\section{Introduction}\label{intro}

Convolutional neural networks~(CNNs) have demonstrated great
capability of solving visual recognition tasks. Since
AlexNet~\cite{krizhevsky2012imagenet} achieved remarkable success on
the ImageNet Challenge~\cite{imagenet_cvpr09}, various deeper and
more complicated
networks~\cite{simonyan2015very,szegedy2015going,he2016deep} have
been proposed to set the performance records. However, the higher
accuracy usually comes with an increasing amount of parameters and
computational cost. For example, the VGG16~\cite{simonyan2015very}
has $128$ million parameters and requires $15,300$ million floating
point operations (FLOPs) to classify an image. In many real-world
applications, predictions need to be performed on resource-limited
platforms such as sensors and mobile phones, thereby requiring
compact models with higher speed. Model compression aims at
exploring a tradeoff between accuracy and efficiency.

Recently, significant progress has been made in the field of model
compression~\cite{iandola2016squeezenet,rastegari2016xnor,wu2016quantized,howard2017mobilenets,zhang2017shufflenet}.
The strategies for building compact and efficient CNNs can be
divided into two categories; those are, compressing pre-trained
networks or designing new compact architectures that are trained
from scratch. Studies in the former category were mostly based on
traditional compression techniques such as product
quantization~\cite{wu2016quantized},
pruning~\cite{see2016compression},
hashing~\cite{chen2015compressing}, Huffman
coding~\cite{han2015deep}, and
factorization~\cite{lebedev2014speeding,jaderberg2014speeding}.

The second category has already been explored before model
compression. Inspired by the Network-In-Network
architecture~\cite{lin2013network},
GoogLeNet~\cite{szegedy2015going} included the Inception module to
build deeper networks without increasing model sizes and
computational cost. Through factorizing convolutions, the Inception
module was further improved by~\cite{szegedy2016rethinking}. The
depth-wise separable convolution, proposed in~\cite{sifre2014rigid},
generalized the factorization idea and decomposed the convolution
into a depth-wise convolution and a $1 \times 1$ convolution. The
operation has been shown to be able to achieve competitive results
with fewer parameters. In terms of model compression,
MobileNets~\cite{howard2017mobilenets} and
ShuffleNets~\cite{zhang2017shufflenet} designed CNNs for mobile
devices by employing depth-wise separable convolutions.

In this work, we focus on the second category and build a new family
of light-weight CNNs known as ChannelNets. By observing that the
fully-connected pattern accounts for most parameters in CNNs, we
propose channel-wise convolutions, which are used to replace dense
connections among feature maps with sparse ones. Early work like
LeNet-5~\cite{LeCun:PIEEE} has shown that sparsely-connected
networks work well when resources are limited. To apply channel-wise
convolutions in model compression, we develop group channel-wise
convolutions, depth-wise separable channel-wise convolutions, and
the convolutional classification layer. They are used to compress
different parts of CNNs, leading to our ChannelNets. ChannelNets
achieve a better trade-off between efficiency and accuracy than
prior compact CNNs, as demonstrated by experimental results on the
ImageNet ILSVRC 2012 dataset. It is worth noting that ChannelNets
are the first models that attempt to compress the fully-connected
classification layer, which accounts for about $25$\% of total
parameters in compact CNNs.

\begin{figure*}[t]
\includegraphics[width=\textwidth]{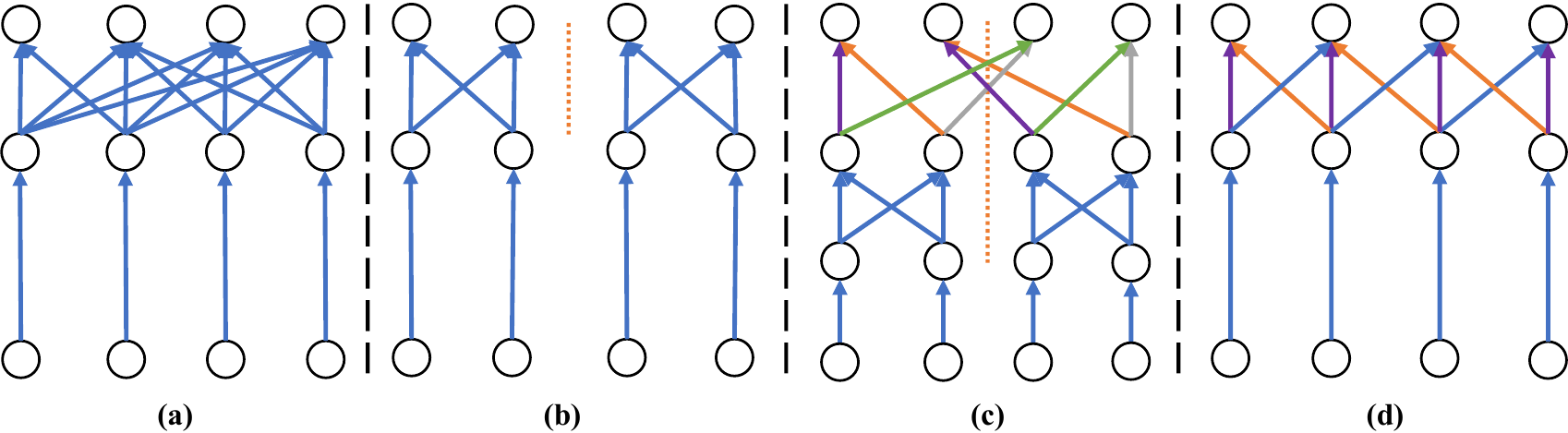}\vspace{-0.3cm}
\caption{Illustrations of different compact convolutions. Part (a)
shows the depth-wise separable convolution, which is composed of a
depth-wise convolution and a $1 \times 1$ convolution. Part (b)
shows the case where the $1 \times 1$ convolution is replaced by a
$1 \times 1$ group convolution. Part (c) illustrates the use of the
proposed group channel-wise convolution for information fusion. Part
(d) shows the proposed depth-wise separable channel-wise
convolution, which consists of a depth-wise convolution and a
channel-wise convolution. For channel-wise convolutions in (c) and
(d), the same color represents shared weights.} \label{fig:ops}
\end{figure*}

\section{Background and Motivations}\label{sec:bg}

The trainable layers of CNNs are commonly composed of convolutional layers and
fully-connected layers. Most prior studies, such as
MobileNets~\cite{howard2017mobilenets} and
ShuffleNets~\cite{zhang2017shufflenet}, focused on compressing convolutional
layers, where most parameters and computation lie. To make the discussion
concrete, suppose a 2-D convolutional operation takes $m$ feature maps with a
spatial size of $d_f \times d_f$ as inputs, and outputs $n$ feature maps of
the same spatial size with appropriate padding. $m$ and $n$ are also known as
the number of input and output channels, respectively. The convolutional
kernel size is $d_k \times d_k$ and the stride is set to $1$. Here, without
loss of generality, we use square feature maps and convolutional kernels for
simplicity. We further assume that there is no bias term in the convolutional
operation, as modern CNNs employ the batch normalization~\cite{ioffe2015batch}
with a bias after the convolution. In this case, the number of parameters in
the convolution is $d_k \times d_k \times m \times n$ and the computational
cost in terms of FLOPs is $d_k \times d_k \times m \times n \times d_f \times
d_f$. Since the convolutional kernel is shared for each spatial location, for
any pair of input and output feature maps, the connections are sparse and
weighted by $d_k \times d_k$ shared parameters. However, the connections among
channels follow a fully-connected pattern, \emph{i.e.}, all $m$ input channels
are connected to all $n$ output channels, which results in the $m \times n$
term. For deep convolutional layers, $m$ and $n$ are usually large numbers
like $512$ and $1024$, thus $m \times n$ is usually very large.

Based on the above insights, one way to reduce the size and cost of
convolutions is to circumvent the multiplication between $d_k \times
d_k$ and $m \times n$. MobileNets~\cite{howard2017mobilenets}
applied this approach to explore compact deep models for mobile
devices. The core operation employed in MobileNets is the depth-wise
separable convolution~\cite{chollet2016xception}, which consists of
a depth-wise convolution and a $1 \times 1$ convolution, as
illustrated in Figure~\ref{fig:ops}(a). The depth-wise convolution
applies a single convolutional kernel independently for each input
feature map, thus generating the same number of output channels. The
following $1 \times 1$ convolution is used to fuse the information
of all output channels using a linear combination. The depth-wise
separable convolution actually decomposes the regular convolution
into a depth-wise convolution step and a channel-wise fuse step.
Through this decomposition, the number of parameters becomes
\begin{equation}
  d_k \times d_k \times m + m \times n,
\label{eq:dwsconv_params}
\end{equation}
and the computational cost becomes
\begin{equation}
  d_k \times d_k \times m \times d_f \times d_f + m \times n \times d_f \times d_f.
\label{eq:dwsconv_cost}
\end{equation}
In both equations, the first term corresponds to the depth-wise
convolution and the second term corresponds to the $1 \times 1$
convolution. By decoupling $d_k \times d_k$ and $m \times n$, the
amounts of parameters and computations are reduced.

While MobileNets successfully employed depth-wise separable
convolutions to perform model compression and achieve competitive
results, it is noted that the $m \times n$ term still dominates the
number of parameters in the models. As pointed out
in~\cite{howard2017mobilenets}, $1 \times 1$ convolutions, which
lead to the $m \times n$ term, account for 74.59\% of total
parameters in MobileNets. The analysis of regular convolutions
reveals that $m \times n$ comes from the fully-connected pattern,
which is also the case in $1 \times 1$ convolutions. To understand
this, first consider the special case where $d_f = 1$. Now the
inputs are $m$ units as each feature map has only one unit. As the
convolutional kernel size is $1 \times 1$, which does not change the
spatial size of feature maps, the outputs are also $n$ units. It is
clear that the operation between the $m$ input units and the $n$
output units is a fully-connected operation with $m \times n$
parameters. When $d_f > 1$, the fully-connected operation is shared
for each spatial location, leading to the $1 \times 1$ convolution.
Hence, the $1 \times 1$ convolution actually outputs a linear
combination of input feature maps. More importantly, in terms of
connections between input and output channels, both the regular
convolution and the depth-wise separable convolution follow the
fully-connected pattern.

As a result, a better strategy to compress convolutions is to change
the dense connection pattern between input and output channels.
Based on the depth-wise separable convolution, it is equivalent to
circumventing the $1 \times 1$ convolution. A simple method,
previously used in AlexNet~\cite{krizhevsky2012imagenet}, is the
group convolution. Specifically, the $m$ input channels are divided
into $g$ mutually exclusive groups. Each group goes through a $1
\times 1$ convolution independently and produces $n/g$ output
feature maps. It follows that there are still $n$ output channels in
total. For simplicity, suppose both $m$ and $n$ are divisible by
$g$. As the $1 \times 1$ convolution for each group requires $1/g^2$
parameters and FLOPs, the total amount after grouping is only $1/g$
as compared to the original $1 \times 1$ convolution.
Figure~\ref{fig:ops}(b) describes a $1 \times 1$ group convolution
where the number of groups is $2$.

However, the grouping operation usually compromises performance
because there is no interaction among groups. As a result,
information of feature maps in different groups is not combined, as
opposed to the original $1 \times 1$ convolution that combines
information of all input channels. To address this limitation,
ShuffleNet~\cite{zhang2017shufflenet} was proposed, where a
shuffling layer was employed after the $1 \times 1$ group
convolution. Through random permutation, the shuffling layer partly
achieves interactions among groups. But any output group accesses
only $m/g$ input feature maps and thus collects partial information.
Due to this reason, ShuffleNet had to employ a deeper architecture
than MobileNets to achieve competitive results.

\section{Channel-Wise Convolutions and ChannelNets}

In this work, we propose channel-wise convolutions in
Section~\ref{sec:CWConv}, based on which we build our ChannelNets.
In Section~\ref{sec:GCWConv}, we apply group channel-wise
convolutions to address the information inconsistency problem caused
by grouping. Afterwards, we generalize our method in
Section~\ref{sec:DWSCWConv}, which leads to a direct replacement of
depth-wise separable convolutions in deeper layers. Through analysis
of the generalized method, we propose a convolutional classification
layer to replace the fully-connected output layer in
Section~\ref{sec:fullyRep}, which further reduces the amounts of
parameters and computations. Finally, Section~\ref{sec:channelnet}
introduces the architecture of our ChannelNets.

\subsection{Channel-Wise Convolutions}\label{sec:CWConv}

We begin with the definition of channel-wise convolutions in
general. As discussed above, the $1 \times 1$ convolution is
equivalent to using a shared fully-connected operation to scan every
$d_f \times d_f$ locations of input feature maps. A channel-wise
convolution employs a shared 1-D convolutional operation, instead of
the fully-connected operation. Consequently, the connection pattern
between input and output channels becomes sparse, where each output
feature map is connected to a part of input feature maps. To be
specific, we again start with the special case where $d_f=1$. The
$m$ input units (feature maps) can be considered as a 1-D feature
map of size $m$. Similarly, the output becomes a 1-D feature map of
size $n$. Note that both the input and output have only $1$ channel.
The channel-wise convolution performs a 1-D convolution with
appropriate padding to map the $m$ units to the $n$ units. In the
cases where $d_f>1$, the same 1-D convolution is computed for every
spatial locations. As a result, the number of parameters in a
channel-wise convolution with a kernel size of $d_c$ is simply $d_c$
and the computational cost is $d_c \times n \times d_f \times d_f$.
By employing sparse connections, we avoid the $m \times n$ term.
Therefore, channel-wise convolutions consume a negligible amount of
computations and can be performed efficiently.

\subsection{Group Channel-Wise Convolutions}\label{sec:GCWConv}

We apply channel-wise convolutions to develop a solution to the information
inconsistency problem incurred by grouping. After the $1 \times 1$ group
convolution, the outputs are $g$ groups, each of which includes $n/g$ feature
maps. As illustrated in Figure~\ref{fig:ops}(b), the $g$ groups are computed
independently from completely separate groups of input feature maps. To
enable interactions among groups, an efficient information fusion layer is
needed after the $1 \times 1$ group convolution. The fusion layer is expected
to retain the grouping for following group convolutions while allowing each
group to collect information from all the groups. Concretely, both inputs and
outputs of this layer should be $n$ feature maps that are divided into $g$
groups. Meanwhile, the $n/g$ output channels in any group should be computed
from all the $n$ input channels. More importantly, the layer must be compact
and efficient; otherwise the advantage of grouping will be compromised.

Based on channel-wise convolutions, we propose the group
channel-wise convolution, which serves elegantly as the fusion
layer. Given $n$ input feature maps that are divided into $g$
groups, this operation performs $g$ independent channel-wise
convolutions. Each channel-wise convolution uses a stride of $g$ and
outputs $n/g$ feature maps with appropriate padding. Note that, in
order to ensure all $n$ input channels are involved in the
computation of any output group of channels, the kernel size of
channel-wise convolutions needs to satisfy $d_c \geq g$. The desired
outputs of the fusion layer is obtained by concatenating the outputs
of these channel-wise convolutions. Figure~\ref{fig:ops}(c) provides
an example of using the group channel-wise convolution after the $1
\times 1$ group convolution, which replaces the original $1 \times
1$ convolution.

To see the efficiency of this approach, the number of parameters of the $1
\times 1$ group convolution followed by the group channel-wise convolution is
$\frac{m}{g} \times \frac{n}{g} \times g + d_c \times g$,
and the computational cost is
$\frac{m}{g} \times \frac{n}{g} \times d_f \times d_f \times g + d_c \times \frac{n}{g} \times d_f \times d_f \times g$.
Since in most cases we have $d_c \ll m$, our approach requires
approximately $1/g$ training parameters and FLOPs, as compared to
the second terms in Eqs.~\ref{eq:dwsconv_params}
and~\ref{eq:dwsconv_cost}.

\subsection{Depth-Wise Separable Channel-Wise Convolutions}\label{sec:DWSCWConv}

Based on the above descriptions, it is worth noting that there is a
special case where the number of groups and the number of input and
output channels are equal, \emph{i.e.}, $g=m=n$. A similar scenario
resulted in the development of depth-wise
convolutions~\cite{howard2017mobilenets,chollet2016xception}. In
this case, there is only one feature map in each group. The $1
\times 1$ group convolution simply scales the convolutional kernels
in the depth-wise convolution. As the batch
normalization~\cite{ioffe2015batch} in each layer already involves a
scaling term, the $1 \times 1$ group convolution becomes redundant
and can be removed. Meanwhile, instead of using $m$ independent
channel-wise convolutions with a stride of $m$ as the fusion layer,
we apply a single channel-wise convolution with a stride of $1$. Due
to the removal of the $1 \times 1$ group convolution, the
channel-wise convolution directly follows the depth-wise
convolution, resulting in the depth-wise separable channel-wise
convolution, as illustrated in Figure~\ref{fig:ops}(d).

In essence, the depth-wise separable channel-wise convolution
replaces the $1 \times 1$ convolution in the depth-wise separable
convolution with the channel-wise convolution. The connections among
channels are changed directly from a dense pattern to a sparse one.
As a result, the number of parameters is $d_k \times d_k \times m + d_c$,
and the cost is
$d_k \times d_k \times m \times d_f \times d_f + d_c \times n \times d_f \times d_f$,
which saves dramatic amounts of parameters and computations. This
layer can be used to directly replace the depth-wise separable
convolution.

\subsection{Convolutional Classification Layer}\label{sec:fullyRep}

\begin{wrapfigure}[20]{r}{0.6\textwidth}\vspace{-0.4cm}
\centering
\includegraphics[width=0.6\textwidth]{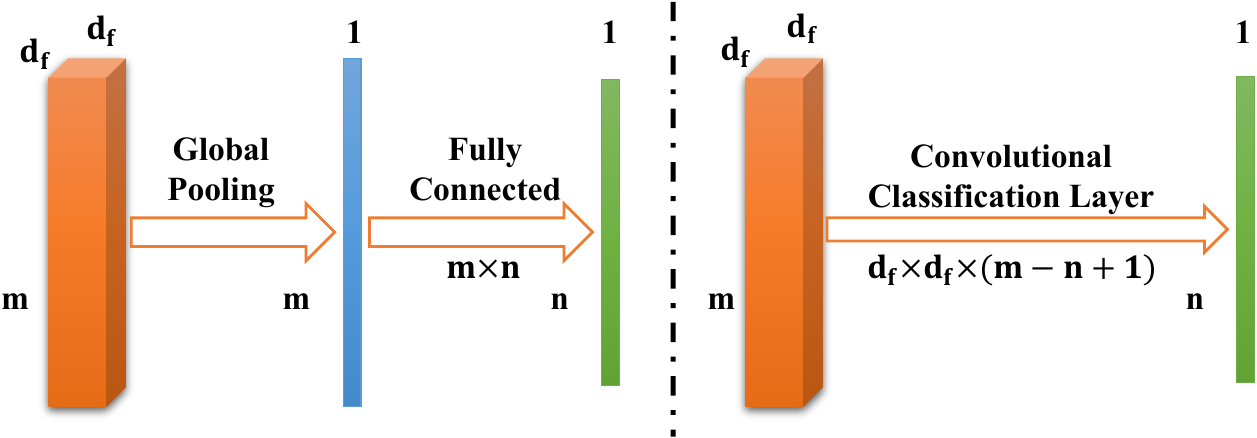}\vspace{-0.3cm}
\caption{An illustration of the convolutional classification layer.
The left part describes the original output layers, \emph{i.e.,} a
global average pooling layer and a fully-connected classification
layer. The global pooling layer reduces the spatial size $d_f \times
d_f$ to $1 \times 1$ while keeping the number of channels. Then the
fully-connected classification layer changes the number of channels
from $m$ to $n$, where $n$ is the number of classes. The right part
illustrates the proposed convolutional classification layer, which
performs a single 3-D convolution with a kernel size of $d_f \times
d_f \times (m-n+1)$ and no padding. The convolutional classification
layer saves a significant amount of parameters and computation.}
\label{fig:fully}
\end{wrapfigure}

Most prior model compression methods pay little attention to the
very last layer of CNNs, which is a fully-connected layer used to
generate classification results. Taking MobileNets on the ImageNet
dataset as an example, this layer uses a $1,024$-component feature
vector as inputs and produces $1,000$ logits corresponding to
$1,000$ classes. Therefore, the number of parameters is $1,024
\times 1,000 \approx 1$ million, which accounts for $24.33$\% of
total parameters as reported in~\cite{howard2017mobilenets}. In this
section, we explore a special application of the depth-wise
separable channel-wise convolution, proposed in
Section~\ref{sec:DWSCWConv}, to reduce the large amount of
parameters in the classification layer.

We note that the second-to-the-last layer is usually a global
average pooling layer, which reduces the spatial size of feature
maps to $1$. For example, in MobileNets, the global average pooling
layer transforms $1,024$ $7 \times 7$ input feature maps into
$1,024$ $1 \times 1$ output feature maps, corresponding to the
$1,024$-component feature vector fed into the classification layer.
In general, suppose the spatial size of input feature maps is $d_f
\times d_f$. The global average pooling layer is equivalent to a
special depth-wise convolution with a kernel size of $d_f \times
d_f$, where the weights in the kernel is fixed to $1/d_f^2$.
Meanwhile, the following fully-connected layer can be considered as
a $1 \times 1$ convolution as the input feature vector can be viewed
as $1 \times 1$ feature maps. Thus, the global average pooling layer
followed by the fully-connected classification layer is a special
depth-wise convolution followed by a $1 \times 1$ convolution,
resulting in a special depth-wise separable convolution.

As the proposed depth-wise separable channel-wise convolution can
directly replace the depth-wise separable convolution, we attempt to
apply the replacement here. Specifically, the same special
depth-wise convolution is employed, but is followed by a
channel-wise convolution with a kernel size of $d_c$ whose number of
output channels is equal to the number of classes. However, we
observe that such an operation can be further combined using a
regular 3-D convolution~\cite{Ji:TPAMI2012}.

In particular, the $m$ $d_f \times d_f$ input feature maps can be
viewed as a single 3-D feature map with a size of $d_f \times d_f
\times m$. The special depth-wise convolution, or equivalently the
global average pooling layer, is essentially a 3-D convolution with
a kernel size of $d_f \times d_f \times 1$, where the weights in the
kernel is fixed to $1/d_f^2$. Moreover, in this view, the
channel-wise convolution is a 3-D convolution with a kernel size of
$1 \times 1 \times d_c$. These two consecutive 3-D convolutions
follow a factorized pattern. As proposed
in~\cite{szegedy2016rethinking}, a $d_k \times d_k$ convolution can
be factorized into two consecutive convolutions with kernel sizes of
$d_k \times 1$ and $1 \times d_k$, respectively. Based on this
factorization, we combine the two 3-D convolutions into a single one
with a kernel size of $d_f \times d_f \times d_c$. Suppose there are
$n$ classes, to ensure that the number of output channels equals to
the number of classes, $d_c$ is set to $(m-n+1)$ with no padding on
the input. This 3-D convolution is used to replace the global
average pooling layer followed by the fully-connected layer, serving
as a convolutional classification layer.

While the convolutional classification layer dramatically reduces
the number of parameters, there is a concern that it may cause a
signification loss in performance. In the fully-connected
classification layer, each prediction is based on the entire feature
vector by taking all features into consideration. In contrast, in
the convolutional classification layer, the prediction of each class
uses only $(m-n+1)$ features. However, our experiments show that the
weight matrix of the fully-connected classification layer is very
sparse, indicating that only a small number of features contribute
to the prediction of a class. Meanwhile, our ChannelNets with the
convolutional classification layer achieve much better results than
other models with similar amounts of parameters.

\subsection{ChannelNets}\label{sec:channelnet}

\begin{wrapfigure}[24]{r}{0.5\textwidth}\vspace{-0.4cm}
\vspace{-5pt}
\includegraphics[width=0.5\textwidth]{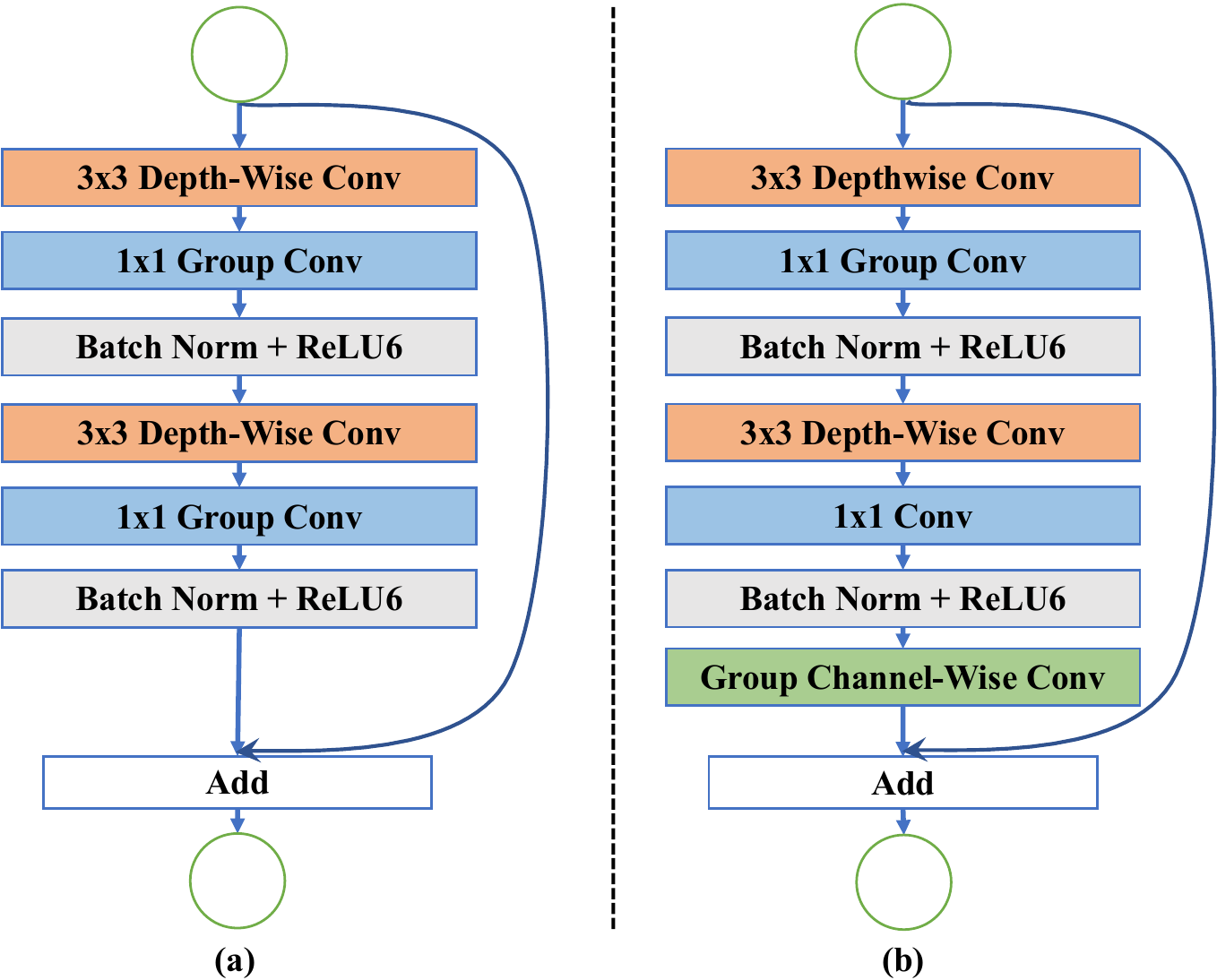}\vspace{-0.3cm}
\caption{Illustrations of the group module~(GM) and the group
channel-wise module (GCWM). Part (a) shows GM, which has two
depth-wise separable convolutional layers. Note that $1 \times 1$
convolutions is replaced by $1 \times 1$ group convolutions to save
computations. A skip connection is added to facilitate model
training. GCWM is described in part (b). Compared to GM, it has a
group channel-wise convolution to fuse information from different
groups.} \label{fig:unit} \vspace{-10pt}
\end{wrapfigure}

With the proposed group channel-wise convolutions, the depth-wise separable
channel-wise convolutions, and the convolutional classification layer, we
build our ChannelNets. We follow the basic architecture of MobileNets to allow
fair comparison and design three ChannelNets with different compression
levels. Notably, our proposed methods are orthogonal to the work of
MobileNetV2~\cite{sandler2018inverted}. Similar to MobileNets, we can apply
our methods to MobileNetV2 to further reduce the parameters and computational
cost. The details of network architectures are shown in
Table~\ref{table:netbody} in the supplementary material.

\textbf{ChannelNet-v1:} To employ the group channel-wise
convolutions, we design two basic modules; those are, the group
module~(GM) and the group channel-wise module~(GCWM). They are
illustrated in Figure~\ref{fig:unit}. GM simply applies $1 \times 1$
group convolution instead of $1 \times 1$ convolution and adds a
residual connection~\cite{he2016deep}. As analyzed above, GM saves
computations but suffers from the information inconsistency problem.
GCWM addresses this limitation by inserting a group channel-wise
convolution after the second $1 \times 1$ group convolution to
achieve information fusion. Either module can be used to replace two
consecutive depth-wise separable convolutional layers in MobileNets.
In our ChannelNet-v1, we choose to replace depth-wise separable
convolutions with larger numbers of input and output channels.
Specifically, six consecutive depth-wise separable convolutional
layers with $512$ input and output channels are replaced by two
GCWMs followed by one GM. In these modules, we set the number of
groups to $2$. The total number of parameters in ChannelNet-v1 is
about $3.7$ million.

\textbf{ChannelNet-v2:} We apply the depth-wise separable
channel-wise convolutions on ChannelNet-v1 to further compress the
network. The last depth-wise separable convolutional layer has $512$
input channels and $1,024$ output channels. We use the depth-wise
separable channel-wise convolution to replace this layer, leading to
ChannelNet-v2. The number of parameters reduced by this replacement
of a single layer is $1$ million, which accounts for about $25$\% of
total parameters in ChannelNet-v1.

\textbf{ChannelNet-v3:} We employ the convolutional classification
layer on ChannelNet-v2 to obtain ChannelNet-v3. For the ImageNet
image classification task, the number of classes is $1,000$, which
means the number of parameters in the fully-connected classification
layer is $1024 \times 1000 \approx 1$ million. Since the number of
parameters for the convolutional classification layer is only $7
\times 7 \times 25 \approx 1$ thousand, ChannelNet-v3 reduces 1
million parameters approximately.

\section{Experimental Studies}

In this section, we evaluate the proposed ChannelNets on the
ImageNet ILSVRC 2012 image classification
dataset~\cite{imagenet_cvpr09}, which has served as the benchmark
for model compression. We compare different versions of ChannelNets
with other compact CNNs. Ablation studies are also conducted to show
the effect of group channel-wise convolutions. In addition, we
perform an experiment to demonstrate the sparsity of weights in the
fully-connected classification layer.

\subsection{Dataset}

The ImageNet ILSVRC 2012 dataset contains $1.2$ million training
images and $50$ thousand validation images. Each image is labeled by
one of $1,000$ classes. We follow the same data augmentation process
in~\cite{he2016deep}. Images are scaled to $256 \times 256$.
Randomly cropped patches with a size of $224 \times 224$ are used
for training. During inference, $224 \times 224$ center crops are
fed into the networks. To compare with other compact
CNNs~\cite{howard2017mobilenets,zhang2017shufflenet}, we train our
models using training images and report accuracies computed on the
validation set, since the labels of test images are not publicly
available.

\subsection{Experimental Setup}

We train our ChannelNets using the same settings as those for MobileNets
except for a minor change. For depth-wise separable convolutions, we remove
the batch normalization and activation function between the depth-wise
convolution and the $1 \times 1$ convolution. We observe that it has no
influence on the performance while accelerating the training speed. For the
proposed GCWMs, the kernel size of group channel-wise convolutions is set to
$8\times 8$. In depth-wise separable channel-wise convolutions, we set the
kernel size to $16\times 16$. In the convolutional classification layer, the
kernel size of the 3-D convolution is $7 \times 7 \times 25$. All models are
trained using the stochastic gradient descent optimizer with a momentum of 0.9
for 80 epochs. The learning rate starts at $0.1$ and decays by $0.1$ at the
45$^{th}$, 60$^{th}$, 65$^{th}$, 70$^{th}$, and 75$^{th}$ epoch.
Dropout~\cite{srivastava2014dropout} with a rate of $0.0001$ is applied after
$1 \times 1$ convolutions. We use 4 TITAN Xp GPUs and a batch size of $512$
for training, which takes about $3$ days.

\subsection{Comparison of ChannelNet-v1 with Other Models}

\begin{wraptable}[14]{r}{0.5\textwidth}\vspace{-0.1cm}
\vspace{-15pt} \centering \caption{Comparison between ChannelNet-v1
and other CNNs in terms of the top-1 accuracy on the ImageNet
validation set, the number of total parameters, and FLOPs needed for
classifying an image.} \label{table:models}
\begin{tabular}{  l   c  c  c }
    \hline
    \textbf{Models} & \textbf{Top-1} & \textbf{Params} & \textbf{FLOPs} \\ \hline\hline
    GoogleNet           & 0.698 & 6.8m & 1550m  \\ \hline
    VGG16               & 0.715 & 128m & 15300m \\ \hline
    AlexNet             & 0.572 & 60m  & 720m   \\ \hline\hline
    SqueezeNet          & 0.575 & 1.3m & 833m   \\ \hline
    1.0 MobileNet       & 0.706 & 4.2m & 569m   \\ \hline
    ShuffleNet 2x       & 0.709 & 5.3m & 524m   \\ \hline
    \textbf{ChannelNet-v1}       & 0.705 & 3.7m & 407m   \\ \hline
\end{tabular}
\vspace{-10pt}
\end{wraptable}

We compare ChannelNet-v1 with other CNNs, including regular networks
and compact ones, in terms of the top-1 accuracy, the number of
parameters and the computational cost in terms of FLOPs. The results
are reported in Table~\ref{table:models}. We can see that
ChannelNet-v1 is the most compact and efficient network, as it
achieves the best trade-off between efficiency and accuracy.

We can see that SqueezeNet~\cite{iandola2016squeezenet} has the
smallest size. However, the speed is even slower than AlexNet and
the accuracy is not competitive to other compact CNNs. By replacing
depth-wise separable convolutions with GMs and GCWMs, ChannelNet-v1
achieves nearly the same performance as $1.0$ MobileNet with a
$11.9$\% reduction in parameters and a $28.5$\% reduction in FLOPs.
Here, the $1.0$ represents the width multiplier in MobileNets, which
is used to control the width of the networks. MobileNets with
different width multipliers are compared with ChannelNets under
similar compression levels in Section~\ref{sec:exper_width}.
ShuffleNet 2x can obtain a slightly better performance. However, it
employs a much deeper network architecture, resulting in even more
parameters and FLOPs than MobileNets. This is because more layers
are required when using shuffling layers to address the information
inconsistency problem in $1 \times 1$ group convolutions. Thus, the
advantage of using group convolutions is compromised. In contrast,
our group channel-wise convolutions can overcome the problem without
more layers, as shown by experiments in
Section~\ref{sec:exper_group}.

\subsection{Comparison of ChannelNets with Models Using Width Multipliers}\label{sec:exper_width}

\begin{wraptable}{r}{0.45\textwidth}\vspace{-0.1cm}
\vspace{-15pt}
\centering \caption{Comparison between ChannelNets and other compact
CNNs with width multipliers in terms of the top-1 accuracy on the
ImageNet validation set, and the number of total parameters. The
numbers before the model names represent width multipliers.}
\label{table:vsmulti}
\begin{tabular}{  l   c  c }
    \hline
    \textbf{Models} & \textbf{Top-1} & \textbf{Params} \\ \hline\hline
    0.75 MobileNet       & 0.684 & 2.6m \\ \hline
    0.75 ChannelNet-v1          & 0.678 & 2.3m \\ \hline
    \textbf{ChannelNet-v2}   & \textbf{0.695} & 2.7m \\ \hline \hline
    0.5 MobileNet    & 0.637 & 1.3m \\ \hline
    0.5 ChannelNet-v1       & 0.627 & 1.2m \\ \hline
    \textbf{ChannelNet-v3}      & \textbf{0.667} & 1.7m \\ \hline
\end{tabular}
\vspace{-10pt}
\end{wraptable}

The width multiplier is proposed in~\cite{howard2017mobilenets} to
make the network architecture thinner by reducing the number of
input and output channels in each layer, thereby increasing the
compression level. This approach simply compresses each layer by the
same factor. Note that most of parameters lie in deep layers of the
model. Hence, reducing widths in shallow layers does not lead to
significant compression, but hinders model performance, since it is
important to maintain the number of channels in the shallow part of
deep models. Our ChannelNets explore a different way to achieve
higher compression levels by replacing the deepest layers in CNNs.
Remarkably, ChannelNet-v3 is the first compact network that attempts
to compress the last layer, \emph{i.e.,} the fully-connected
classification layer.

We perform experiments to compare ChannelNet-v2 and ChannelNet-v3
with compact CNNs using width multipliers. The results are shown in
Table~\ref{table:vsmulti}. We apply width multipliers $\{0.75,
0.5\}$ on both MobileNet and ChannelNet-v1 to illustrate the impact
of applying width multipliers. In order to make the comparison fair,
compact networks with similar compression levels are compared
together. Specifically, we compare ChannelNet-v2 with 0.75 MobileNet
and 0.75 ChannelNet-v1, since the numbers of total parameters are in
the same 2.x million level. For ChannelNet-v3, 0.5 MobileNet and 0.5
ChannelNet-v1 are used for comparison, as all of them contain 1.x
million parameters.

We can observe from the results that ChannelNet-v2 outperforms
$0.75$ MobileNet with an absolute $1.1$\% gain in accuracy, which
demonstrates the effect of our depth-wise separable channel-wise
convolutions. In addition, note that using depth-wise separable
channel-wise convolutions to replace depth-wise separable
convolutions is a more flexible way than applying width multipliers.
It only affects one layer, as opposed to all layers in the networks.
ChannelNet-v3 has significantly better performance than $0.5$
MobileNet by $3$\% in accuracy. It shows that our convolutional
classification layer can retain the accuracy to most extent while
increasing the compression level. The results also show that
applying width multipliers on ChannelNet-v1 leads to poor
performance.

\subsection{Ablation Study on Group Channel-Wise Convolutions}\label{sec:exper_group}

\begin{wraptable}{r}{0.42\textwidth}
\vspace{-17pt} \centering \caption{Comparison between ChannelNet-v1
and ChannelNet-v1 without group channel-wise convolutions, denoted
as ChannelNet-v1(-). The comparison is in terms of the top-1
accuracy on the ImageNet validation set, and the number of total
parameters.} \label{table:abla}
\begin{tabular}{  l   c  c }
    \hline
    \textbf{Models} & \textbf{Top-1} & \textbf{Params} \\ \hline\hline
    ChannelNet-v1(-)        & 0.697 & 3.7m \\ \hline
    ChannelNet-v1           & 0.705 & 3.7m \\ \hline
\end{tabular}
\vspace{-10pt}
\end{wraptable}

To demonstrate the effect of our group channel-wise convolutions, we
conduct an ablation study on ChannelNet-v1. Based on ChannelNet-v1, we replace
the two GCWMs with GMs, thereby removing all group channel-wise convolutions.
The model is denoted as ChannelNet-v1(-). It follows exactly the same
experimental setup as ChannelNet-v1 to ensure fairness. Table~\ref{table:abla}
provides comparison results between ChannelNet-v1(-) and ChannelNet-v1.
ChannelNet-v1 outperforms ChannelNet-v1(-) by $0.8$\%, which is
significant as ChannelNet-v1 has only $32$ more parameters with group
channel-wise convolutions. Therefore, group channel-wise convolutions are
extremely efficient and effective information fusion layers for solving the
problem incurred by group convolutions.

\subsection{Sparsity of Weights in Fully-Connected Classification Layers}

In ChannelNet-v3, we replace the fully-connected classification layer with our
convolutional classification layer. Each prediction is based on only
$(m-n+1)$ features instead of all $n$ features, which raises a concern of
potential loss in performance. To investigate this further, we analyze the
weight matrix in the fully-connected classification layer, as shown in
Figure~\ref{fig:fully_view} in the supplementary material. We take the fully-
connected classification layer of ChannelNet-v1 as an example. The analysis
shows that the weights are sparsely distributed in the weight matrix, which
indicates that each prediction only makes use of a small number of features,
even with the fully-connected classification layer. Based on this insight, we
propose the convolutional classification layer and ChannelNet-v3. As shown in
Section~\ref{sec:exper_width}, ChannelNet-v3 is highly compact and efficient
with promising performance.

\section{Conclusion and Future Work}

In this work, we propose channel-wise convolutions to perform model
compression by replacing dense connections in deep networks. We
build a new family of compact and efficient CNNs, known as
ChannelNets, by using three instances of channel-wise convolutions;
namely group channel-wise convolutions, depth-wise separable
channel-wise convolutions, and the convolutional classification
layer. Group channel-wise convolutions are used together with $1
\times 1$ group convolutions as information fusion layers.
Depth-wise separable channel-wise convolutions can be directly used
to replace depth-wise separable convolutions. The convolutional
classification layer is the first attempt in the field of model
compression to compress the the fully-connected classification
layer. Compared to prior methods, ChannelNets achieve a better
trade-off between efficiency and accuracy. The current study
evaluates the proposed methods on image classification tasks, but
the methods can be applied to other tasks, such as detection and
segmentation. We plan to explore these applications in the future.




\clearpage

\bibliographystyle{plainnat}
{\scriptsize \bibliography{deep}}

\clearpage

\center{{\LARGE{Supplementary Material for ``ChannelNets: Compact
and Efficient Convolutional Neural Networks via Channel-Wise
Convolutions''}}

\begin{table}[h]
\centering \caption{ChannelNets Architectures. The operations are
described in the format of ``Type / Stride / \# Output channels''.
``Conv'' denotes the regular convolution; ``DWSConv'' denotes the
depth-wise separable convolution; ``DWSCWConv'' denotes the
depth-wise separable channel-wise convolution; ``CCL'' denotes the
convolutional classification layer. The kernel size in regular
convolutions and depth-wise convolutions is $3 \times 3$.}
\label{table:netbody}
\begin{tabular}{  c  | c  | c }
    \hline
    \textbf{v1} & \textbf{v2} & \textbf{v3} \\ \hline\hline
    \multicolumn{3}{c}{Conv   / 2 / 32}       \\ \hline
    \multicolumn{3}{c}{DWSConv / 1 / 64}      \\ \hline
    \multicolumn{3}{c}{DWSConv / 2 / 128}     \\ \hline
    \multicolumn{3}{c}{DWSConv / 1 / 128}    \\ \hline
    \multicolumn{3}{c}{DWSConv / 2 / 256}    \\ \hline
    \multicolumn{3}{c}{DWSConv / 1 / 256}    \\ \hline
    \multicolumn{3}{c}{DWSConv / 2 / 512}    \\ \hline
    \multicolumn{3}{c}{GCWM / 1 / 512}       \\ \hline
    \multicolumn{3}{c}{GCWM / 1 / 512}      \\ \hline
    \multicolumn{3}{c}{GM / 1 / 512}      \\ \hline
    \multicolumn{3}{c}{DWSConv / 2 / 1024}   \\ \hline
    DWSConv / 1 / 1024     & DWSCWConv & DWSCWConv \\ \hline
    AvgPool + FC                      & AvgPool + FC  & CCL \\
    \hline
\end{tabular}
\end{table}

\begin{figure*}[h]
\centering \includegraphics[width=0.9\textwidth]{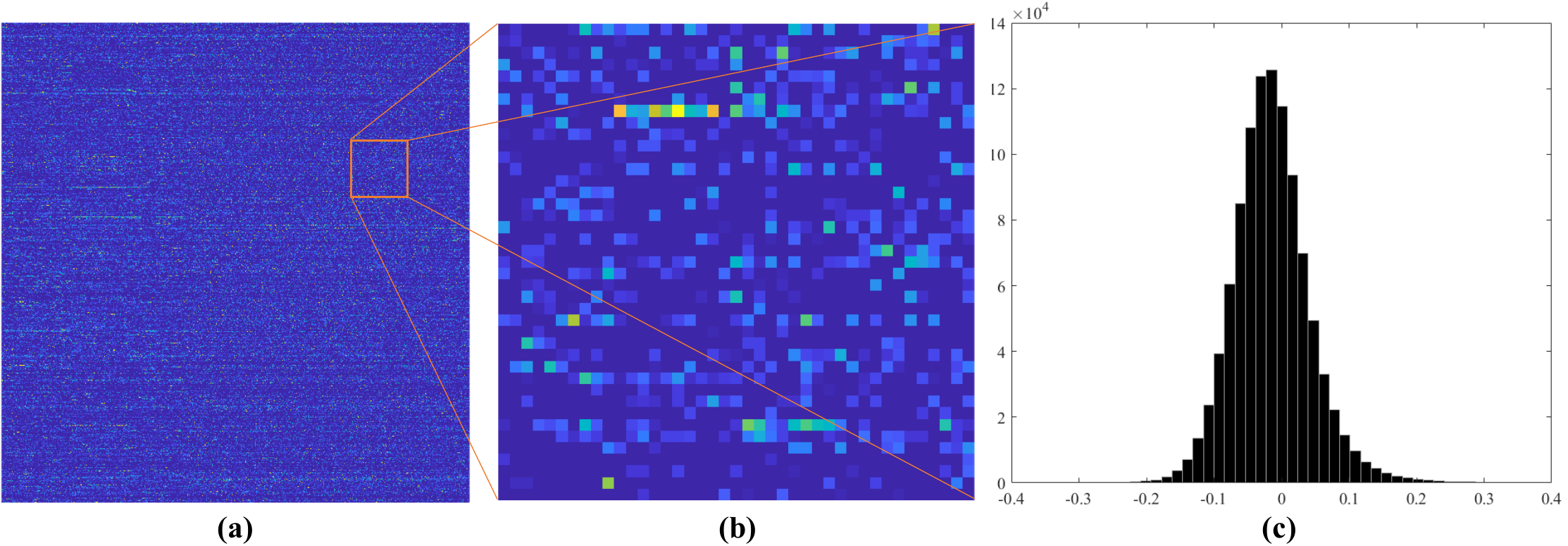}
\caption{An example of the weight patterns in the fully-connected
classification layer of ChannelNet-v1. Part (a) shows the weight
matrix of the fully-connected classification layer. We can see that
the weights are sparsely distributed, as most of the weights are in
blue color, which indicates zero or near zero values. Part (b) gives
a close look of weights in a small region. The histogram in part (c)
statistically demonstrates the sparsity of weights.}
\label{fig:fully_view}
\end{figure*}

\end{document}